# Computational Advantages of Relevance Reasoning in Bayesian Belief Networks

**Yan Lin**
University of Pittsburgh
Intelligent Systems Program
Pittsburgh, PA 15260
*yan@isp.pitt.edu*

**Marek J. Druzdzel**
University of Pittsburgh
Department of Information Science
and Intelligent Systems Program
Pittsburgh, PA 15260
*marek@sis.pitt.edu*

## Abstract

This paper introduces a computational framework for reasoning in Bayesian belief networks that derives significant advantages from focused inference and relevance reasoning. This framework is based on $d$-separation and other simple and computationally efficient techniques for pruning irrelevant parts of a network. Our main contribution is a technique that we call *relevance-based decomposition*. Relevance-based decomposition approaches belief updating in large networks by focusing on their parts and decomposing them into partially overlapping subnetworks. This makes reasoning in some intractable networks possible and, in addition, often results in significant speedup, as the total time taken to update all subnetworks is in practice often considerably less than the time taken to update the network as a whole. We report results of empirical tests that demonstrate practical significance of our approach.

## 1 Introduction

Emergence of probabilistic graphs, such as Bayesian belief networks (BBNs) [Pearl, 1988] and closely related influence diagrams [Shachter, 1986] has made it possible to base uncertain inference in knowledge-based systems on the sound foundations of probability theory and decision theory. Probabilistic graphs offer an attractive knowledge representation tool for reasoning in knowledge-based systems in the presence of uncertainty, cost, preferences, and decisions. They have been successfully applied in such domains as diagnosis, planning, learning, vision, and natural language processing.[1] As many practical systems tend to be large, the main problem faced by the decision-theoretic approach is the complexity of probabilistic reasoning, shown to be NP-hard both for exact inference [Cooper, 1990] and for approximate [Dagum and Luby, 1993] inference.

The critical factor in exact inference schemes is the topology of the underlying graph and, more specifically, its connectivity. The complexity of approximate schemes may, in addition, depend on factors like the a-priori likelihood of the observed evidence or asymmetries in probability distributions. There are a number of ingeniously efficient algorithms that allow for fast belief updating in moderately sized models.[2] Still, each of them is subject to the growth in complexity that is generally exponential in the size of the model. Given the promise of the decision-theoretic approach and an increasing number of its practical applications, it is important to develop schemes that will reduce the computational complexity of inference. Even though the worst case will remain NP-hard, many practical cases may become tractable by, for example, exploring the properties of practical models, approximating the inference, focusing on smaller elements of the models, reducing the connectivity of the underlying graph, or by improvements in the inference algorithms that reduce the constant factor in the otherwise exponential complexity.

In this paper, we introduce a computational framework for reasoning in Bayesian belief networks that derives significant advantages from focused inference and relevance reasoning. We introduce a technique called *relevance-based decomposition*, that computes the marginal distributions over variables of interest by decomposing a network into partially overlapping sub-networks and performing the computation in the identified sub-networks. As this procedure is able, for most reasonably sparse topologies, to identify sub-networks that are significantly smaller than the entire model, it also can be used to make computation in large networks doable, under practical limitations of the available hardware. In addition, we demonstrate empirically that it can lead to significant speedups in large practical models for the clus-tering algorithm [Lauritzen and Spiegelhalter, 1988, Jensen *et al.*, 1990].

---

[1]Some examples of real-world applications are described in a special issue of *Communications of the ACM*, on practical applications of decision-theoretic methods in AI, Vol. 38, No. 3, March 1995.

[2]For an overview of various exact and approximate approaches to algorithms in BBNs see [Henrion, 1990].

All random variables used in this paper are multiple-valued, discrete variables. Lower case letters (e.g., $x$) will represent random variables, and indexed lower-case letters (e.g., $x_i$) will denote their outcomes. In case of binary random variables, the two outcomes will be denoted by upper case (e.g., the two outcomes of a variable $c$ will be denoted by $C$ and $\overline{C}$).

The remainder of this paper is structured as follows. Section 2 reviews briefly the methods of relevance reasoning applied in our framework. Section 3 discusses in somewhat more depth an important element of relevance reasoning, nuisance node removal. Nuisance node removal is the prime tool for significant reductions of clique sizes when clustering algorithms are subsequently applied. Section 4 discusses relevance-based decomposition. Section 5 presents empirical results. Finally, Section 6 discusses the impact of our results on the work on belief updating algorithms for Bayesian belief networks.

## 2    Relevance Reasoning in Bayesian Belief Networks

The concept of relevance is relative to the model, to the focus of reasoning, and to the context in which reasoning takes place [Druzdzel and Suermondt, 1994]. The focus is normally a set of variables of interest $\mathcal{T}$ ($\mathcal{T}$ stands for the *target* variables) and the context is provided by observing the values of some subset $\mathcal{E}$ ($\mathcal{E}$ stands for the *evidence* variables) of other variables in the model.

The cornerstone of most relevance-based methods is probabilistic independence, captured in graphical models by a condition known as $d$–separation [Pearl, 1988], which ties the concept of conditional independence to the structure of the graph. Informally, an evidence node blocks the propagation of information from its ancestors to its descendants, but it also makes all its ancestors interdependent. In this section, we will give the flavor of simple algorithms for relevance-based reasoning in graphical models that we applied in our framework.

Parts of the model that are probabilistically independent from the target nodes $\mathcal{T}$ given the observed evidence $\mathcal{E}$ are computationally irrelevant to reasoning about $\mathcal{T}$. Geiger *et al.* [1990b] show an efficient algorithm for identifying nodes that are probabilistically independent from a set of target nodes given a set of evidence nodes. Removing such nodes can lead to significant savings in computation. Figure 1–a presents a sample network reproduced from Lauritzen and Spiegelhalter [1988]. For example, node $a$ is independent of node $f$ if neither $c$, $d$, or $h$ are observed (Figure 1–a). If nodes $f$ and $d$ are observed, node $g$ will become independent of nodes $a$, $b$, $c$, and $e$, but nodes $b$ and $e$ will become dependent.

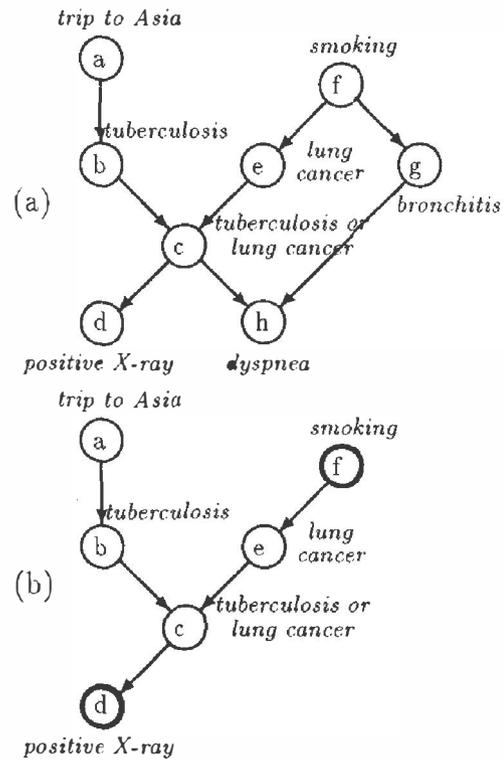

Figure 1: An example of relevance reasoning: removal of nodes based on the $d$–separation condition and barren nodes. If $\mathcal{E} = \{d, f\}$ is the set of evidence nodes and $\mathcal{T} = \{b, e\}$ is the set of target nodes, then nodes $h$ and $g$ are barren.

The next step in reducing the graph is removal of *barren nodes* [Shachter, 1986]. Nodes are barren if they are neither evidence nodes nor target nodes and they have no descendants or if all their descendants are barren. Barren nodes may depend on the evidence, but they do not contribute to the change in probability of the target nodes and are, therefore, computationally irrelevant. A simple extension to the algorithm for identifying independence can remove all barren nodes efficiently [Geiger *et al.*, 1990b, Baker and Boult, 1991]. Figure 1 illustrates the construction of a relevant sub-network from the original network that is based on the $d$–separation criterion. Starting with the network in Figure 1–a, a set of evidence nodes $\mathcal{E} = \{d, f\}$, and a set of target nodes $\mathcal{T} = \{b, e\}$, we obtain the network in Figure 1–b by removing barren nodes $g$ and $h$. (Once node $h$ removed, we can also view node $g$ as $d$–separated from $\mathcal{T}$ by the evidence $\mathcal{E}$.) Networks (a) and (b) are equivalent with respect to computing the posterior probabilities of $\mathcal{T}$ given $\mathcal{E}$.

Schemes based on $d$–separation can be further enhanced by exploration of independences encoded implicitly in conditional probability distributions, including context-specific independences. Some examples of such independences are listed by Druzdzel and Suermondt [1994]. Other relevant work is by Boutilier *et al.* [1996], Heckerman [1990], Heckerman and Breese [1994], Smith *et al.* [1993], and Poole [1993].

The above example illustrates that relevance reasoning can yield smaller sub-networks that are much smaller and less densely connected than the original network. The network in Figure 1–b, in particular, is singly connected and can be solved in polynomial time. This can lead to dramatic improvements in performance, as most relevance algorithms operate on the structural properties of graphs and their complexity is polynomial in the number of arcs in the network (see Druzdzel and Suermondt [1994] for a brief review of relevance-based algorithms).

There are two additional simple methods that we implemented in our framework. The first method, termed evidence propagation, consists of instantiating nodes in the network if their values are indirectly implied by the evidence. The observed evidence may be causally sufficient to imply the values of other, as yet unobserved nodes (e.g., if a patient is male, it implies that he is not pregnant). Similarly, observed evidence may imply other nodes that are causally necessary for that evidence to occur (e.g., observing that a car starts implies that the battery is not empty). Each instantiation reduces the number of uncertain variables and, hence, the computational complexity of inference. Further, instantiations can lead to additional reductions, as they may screen off other variables by making them independent of the variables of interest.

The second method involves absorbing instantiated nodes into the probability distributions of their children. Once we know the state of an observed node, the probabilities of all other states becomes zero and there is no need to store distributions which depend upon those states in its successors. We can modify the probability distribution of its successors and remove the arcs between them. The practical significance of this operation is that the conditional probability tables become smaller and this reduces both the memory and computational requirements. Evidence absorption is closely related to the operation by that name in the Lauritzen and Spiegelhalter's [1988] clustering algorithm and has been studied in detail by Shachter [1990].

We should remark here that in cases where all nodes belong to the target set $\mathcal{T}$, most of the techniques reviewed in this section cannot reduce the size of any cliques in the network — since everything is relevant, nothing can be removed. Evidence absorption, however, removes all outgoing arcs of the evidence nodes and, thereby, reduces the size of some cliques and guarantees to produce less complex networks, unless all evidence nodes are leaf nodes. Of course, the clustering algorithms can be improved to reduce the clique size in practice, but this reduction usually amounts to reduction in computation and not in memory size taken by a clique, as it is done after the network has been compiled into a clique tree. The evidence absorption scheme achieves such reduction before constructing the junction tree. Lastly, we want to point out that evidence absorption often results in more removal of nuisance nodes, which is the subject of the next section.

## 3    Nuisance Nodes

Druzdzel and Suermondt [1994] introduced a class of nodes called *nuisance nodes* and emphasized that they are also reducible by relevance reasoning. Nuisance nodes consist of those predecessor nodes that do not take active part in propagation of belief from the evidence to the target.

Before discussing removal of nuisance nodes, we will define them formally — we believe that this might help to avoid misunderstanding. Of the definitions below, trail, head-to-head node, and active trail are based on Geiger *et al.* [1990a].

**Definition 1 (trail in undirected graph)** *A* trail *in an undirected graph is an alternating sequence of nodes and arcs of the graph such that every arc joins the nodes immediately preceding it and following it.*

**Definition 2 (trail)** *A* trail *in a directed acyclic graph is an alternating sequence of arcs and nodes of the graph that form a trail in the underlying undirected graph.*

**Definition 3 (head-to-head node)** *A node $c$ is called a* head-to-head *node with respect to a trail $t$ if there are two consecutive arcs $a \rightarrow c$ and $c \leftarrow b$ on $t$.*

**Definition 4 (minimal trail)** *A trail connecting $a$ and $b$ in which no node appears more than once is called a* minimal trail *between $a$ and $b$.*

**Definition 5 (active trail)** *A trail $t$ connecting nodes $a$ and $b$ is said to be* active *given a set of nodes $\mathcal{E}$ if (1) every head-to-head node with respect to $t$ either is in $\mathcal{E}$ or has a descendant in $\mathcal{E}$ and (2) every other node on $t$ is outside $\mathcal{E}$.*

**Definition 6 (evidential trail)** *A minimal active trail between an evidence node $e$ and a node $n$, given a set of nodes $\mathcal{E}$, is called an* evidential trail *from $e$ to $n$ given $\mathcal{E}$.*

In case of reducing a network for the sake of explanation of reasoning, the original application of nuisance nodes, the assumption was that only the evidential trails from $\mathcal{E}$ to $\mathcal{T}$ are relevant for explaining the impact of $\mathcal{E}$ on $\mathcal{T}$. Nuisance node is defined with respect to $\mathcal{T}$, $\mathcal{E}$, and all evidential trails between them.

**Definition 7 (nuisance node)** *A* nuisance *node, given evidence $\mathcal{E}$ and target $\mathcal{T}$, is a node that is computationally related to $\mathcal{T}$ given $\mathcal{E}$ but is not part of any evidential trail from any node in $\mathcal{E}$ to any node in $\mathcal{T}$.*

Nuisance nodes are computationally related because they are ancestors of some nodes on a $d$–connecting path (please, note that they cannot be $d$–separated or barren, as they have to be computationally related). We will introduce the concept of *nuisance anchor* defined as follows:

**Definition 8 (nuisance anchor)** *A* nuisance anchor *is a node on an evidential trail that has at least one immediate predecessor that is a nuisance node.*

We will aim to remove entire groups of connected nuisance nodes, which will be captured by the following two definitions:

**Definition 9 (nuisance graph)** *A* nuisance graph *is a subgraph consisting of an anchor and all its nuisance ancestors.*

**Definition 10 (nuisance tree)** *A* nuisance tree *is a nuisance graph that is a polytree.*

Since no barren nodes exist in a network that contains only computationally related nodes, it is a straightforward process to demonstrate that nuisance graphs consist of only ancestors of nuisance anchors.

Finally, it is convenient for the sake of explanation to define the concept of bold nuisance nodes:

**Definition 11 (bold nuisance node)** *A* nuisance node *is called* bold *if it has no ancestors.*

The definition of nuisance nodes provides a straightforward criterion for identifying them in a graphical model. Identification of nuisance graphs can be performed by a variant of the Depth-First-Search algorithm that has complexity O($e$), where $e$ is the number of arcs in the network. The algorithm in Figure 2 for identifying nuisance nodes in directed acyclic graphs is a revised version of the non-separable component algorithm [Even, 1979]. Since all descendant nodes of target or evidence in the (pruned) computational relevant subnetwork can not be nuisance nodes, we mark them $ACTIVE$ first. Then following an arc from an active node to its parent, we find a non-separable component, which is a nuisance graph if it does not contain any active nodes.

To marginalize a nuisance graph into its anchor we need to know the joint probability distribution of those nodes in the graph that are the anchor's parents. In case the graph is a tree, the parents are independent and the tree can be reduced by a recursive marginalization of its bold nuisance nodes until the entire nuisance tree is reduced.

Suppose (Figure 3–a) that the evidence set is $\mathcal{E} = \{d\}$ and the target set is $\mathcal{T} = \{e\}$. Nodes $a$ and $b$ form a nuisance tree with anchor at $c$ and node $f$ forms a one-node nuisance tree with anchor in $c$. Nuisance nodes $a$ and $f$ are bold. In order to reduce both trees into their anchors, we need to successively marginalize their bold

```
Given: A computationally relevant
  Bayesian Network net:
      a set of target nodes T.
      a set of evidence nodes E.
void Mark_Nuisance_Nodes(net)
  empty stack s;
  for each node n in the network
  do n.k := ●
      if (n is a descendant of
         target or evidence)
      then n.mark := ACTIVE
      else n.mark := CLEAN
  for each arc in the network
  do arc.mark := UNVISITED
while (there is still an ACTIVE nodes n
      that has UNVISITED incident
      arcs to its parents)    do
  v := n;   v.f := nil; push v to stack s;
  i := 1;   v.k := i;   v.l := i;
  repeat
    while (v has UNVISITED incident
           arc)    do
      follow arc to find the node u
      arc.mark := VISITED
      if u.k = 0 then
        if u.k < v.l then v.l := u.k;
      else u.f := v;   v := u;
           push v to stack s;
           i := i + 1;   v.k := i;   l.v := i
    end
    if (v.f.k = 1 or v.l >= v.f.k) then
      pop all nodes from stack s down to
      (including) v;
      these nodes with v.f forms a
      non-separable set.
      if (no ACTIVE nodes in this set)
      then mark all nodes in this set
        NUISANCE
      else mark all nodes in this set
        ACTIVE
    else if (v.l < v.f.l) then v.f.l := v.l
    v := v.f;
  until (v.f = nil or v has no
         UNVISITED incident arc)
end
```

Figure 2:  The algorithm for identifying nuisance nodes.

nodes into their descendants, $f$ into $e$, $a$ into $b$, and finally $b$ into $c$. The last operation, in particular, is performed using the following formulas:

$$\Pr(C|E) = \Pr(C|E, \overline{B}) \Pr(\overline{B}) + \Pr(C|E, B) \Pr(B)$$
$$\Pr(C|\overline{E}) = \Pr(C|\overline{E}, \overline{B}) \Pr(\overline{B}) + \Pr(C|\overline{E}, B) \Pr(B)$$

An operation that is analogous to nuisance node removal in networks consisting of Noisy-OR nodes [Pearl, 1988], is also performed by the Netview program described by Pradhan, *et al.* [Pradhan *et al.*, 1994].

Marginalization of nuisance graphs that are not nuisance trees is less straightforward: to be able to remove a nuisance graph, we need to first construct a new conditional probability table for the nuisance anchor. Temporarily forgetting about the evidence present in the network, we condition on the non-nuisance parents of the nuisance anchor and treat the nuisance anchor as a target node. The rest of the network below the nuisance anchor is computationally irrelevant to the target. Any standard inference algorithm can be used to compute the conditional probability distribution of the nuisance anchor, which can then be used to merge the entire nuisance graph into the anchor. When all parents of the nuisance anchor themselves are nuisance nodes, we can remove the entire nuisance graph by computing the prior probability distribution of the nuisance anchor. While computing the probability distribution over a nuisance anchor is hard in general, this probability distribution can be precomputed in advance for some of the those subnetworks that are potential nuisance graphs. Please, note that the probability distribution over a potential noise anchors is not conditioned on any evidence, which makes such precomputation feasible.

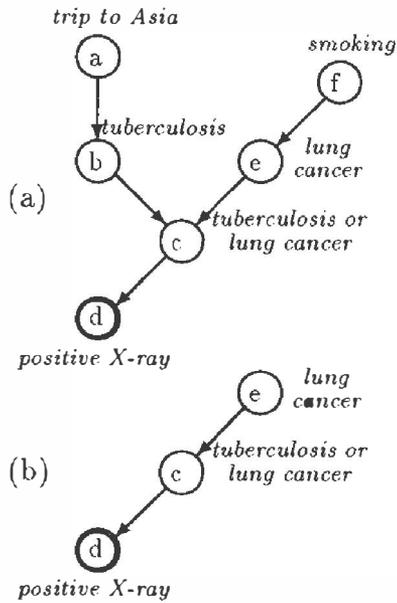

Figure 3: Removal of nuisance nodes $a$, $b$, and $f$.

Removal of a nuisance tree originating from a nuisance anchor reduces one dimension of the conditional probability table (and hence, clique) containing the nuisance anchor and its remaining parents. In the case of nuisance graphs that are not trees, while their removal may lead to significant computational advantages, the computation related to establishing their marginal probability may be in itself complex. To make the marginalization of nuisance graphs worthwhile, it is possible to cache conditional probability tables for those cases that are commonly encountered. These cache tables can be computed at the time the model is constructed and stored for the efficiency of later reasoning. Please note that tables only at those nodes that are potential anchors (identified by the algorithm of Figure 2) need to be precomputed and stored.

## 4   Relevance-Based Decomposition

It is quite obvious that relevance reasoning can lead to significant computational savings if the reasoning is focused, i.e., if the user is interested only in a subset of nodes in the network. (We report almost three orders of magnitude improvement in a very large medical diagnostic network in Section 5.) Relevance-based methods can be very useful even if no target nodes are specified, i.e., when all nodes in a network are of interest. When the original network is large, computing the posterior distribution over all nodes may become intractable: for example, due to excessive memory requirements of the clustering algorithm. In such cases, we can attempt to divide the network into several partially overlapping subnetworks, where all sets combined cover the entire network. Focusing on each of these small subnetworks in separation leads eventually to updating the beliefs of all nodes in the network.

The main problem is, of course, dividing the network. We accomplish that by choosing at each step $i$ a small set of target variables $\mathcal{T}_i$ and pruning those nodes in the network that are not computationally relevant to updating the probability of $\mathcal{T}_i$ given $\mathcal{E}$. Since not all nodes in the network are computationally relevant to $\mathcal{T}_i$, the size of relevant subnetworks can be much smaller than that of the original network. The order in which the target sets $\mathcal{T}_i$ are selected is crucial for the performance of the algorithm. Obviously, with a wrong choice of $\mathcal{T}_i$, the subsets may overlap too much and lead to performance deterioration. Useful heuristics that will minimize the overlap among various subnetworks remain still to be studied. We have observed, however, that even with a very crude choice of the target sets $\mathcal{T}_i$, not only can we handle many intractable networks, but also decrease the total computation time in tractable networks. (We report four-fold increase in speed in Section 5.) This, of course, is not guaranteed and depends on the topology of the network. In very densely connected networks, everything may be relevant to everything, no matter what target set we choose. Such networks, however, would be intractable for any exact inference algorithm.

A sketch of the algorithm outlined informally above is given in Figure 4. We choose at each step $i$ a set of target nodes $\mathcal{T}_i$. Subsequently, we use the relevance reasoning techniques outlined in Sections 2 and 3 to identify a subnetwork that is relevant for computing the posterior probability of $\mathcal{T}_i$ given $\mathcal{E}$. Finally, we employ a standard inference algorithm to compute the posterior probability distribution over the target nodes. Since in general the identified subnetwork will include other nodes than $\mathcal{T}_i$ and $\mathcal{E}$, we update a part of the network. We proceed by focusing on different network nodes from among those that have not yet been up-

```
Given: A Bayesian belief network net,
       a set of evidence nodes E,
void
Relevance_Based_Decomposition
       (net, E)
    while there are still nodes that
          need updating
       Choose a set of target nodes T_i from
          among those that need updating;
       Identify the set S_i of nodes that are
          relevant to computing the
          posterior probability of T_i
          given the set of evidence nodes E;
       Perform belief updating on S_i;
    end
end
```

Figure 4: A basic algorithm for relevance-based decomposition.

dated until all nodes have been updated.

Figure 5 shows a simple example of relevance-based decomposition, given evidence node $\mathcal{E} = \{d\}$ and the choice of targets in different steps: $\mathcal{T}_1 = \{a\}$, $\mathcal{T}_2 = \{g\}$, and $\mathcal{T}_3 = \{h\}$. We decompose the network into three subnetworks. Please note that network $S_2$ is a subset of $S_3$, which leads to redundant computation. We could avoid this by choosing $h$ as a target before choosing $g$.

## 5   Empirical Results

In this section, we present the results of an empirical test of our relevance-based framework for Bayesian belief network inference. We focused our tests on the most surprising result: impact of relevance-based network decomposition on the computational complexity ●f the inference. The algorithm that we used in all tests is an efficient implementation of the clustering algorithm that was made available to us by Alex Kozlov. See Kozlov and Singh [1996] for details of the implementation and some benchmarks. We have enhanced Kozlov's implementation with relevance techniques described in this paper. We have not included caching the probability distributions of nuisance anchors in our tests.

We tested our algorithms using the CPCS network, a multiply-connected multi-layer network consisting of 422 multi-valued nodes and covering a subset of the domain of internal medicine [Pradhan et al., 1994]. Among the 422 nodes, 14 nodes describe diseases, 33 nodes describe history and risk factors, and the remaining 375 nodes describe various findings related to the diseases. The CPCS network is among the largest real networks available to the research community at present time.

Our computer (a Sun Ultra-2 workstation with two 168Mhz UltraSPARC-1 CPU's, each CPU has a 0.5MB L2 cache, the total system RAM memory of

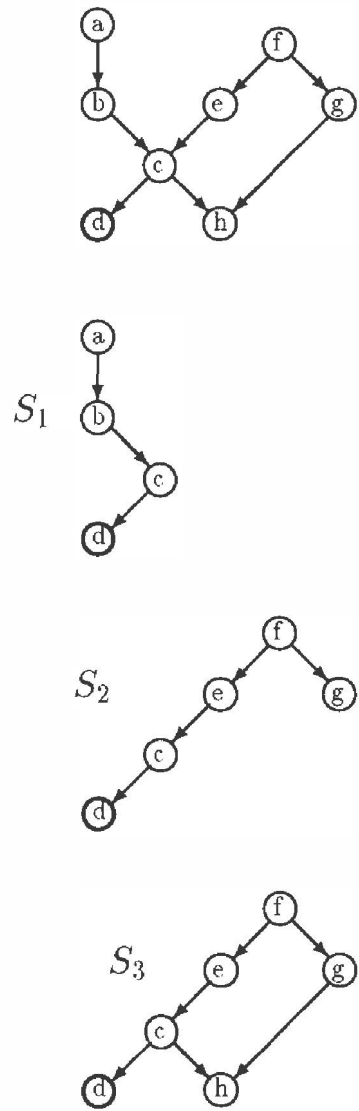

Figure 5: An example of relevance-based decomposition: Given the evidence node $\mathcal{E} = \{d\}$ and the targets $\mathcal{T}_1 = \{a\}$, $\mathcal{T}_2 = \{g\}$, and $\mathcal{T}_3 = \{h\}$, we obtain at each step simple, smaller networks ($S_1$, $S_2$, and $S_3$).

384 MB) was unable to load, compile, and store the entire network in memory and we decided to use a subset consisting of 360 nodes generated by Alex Kozlov for earlier benchmarks of his algorithm. This network is a subset of the full 422 node CPCS network without predisposing factors (like gender, age, smoking, etc.). This reduction is realistic, as history nodes can usually be instantiated and absorbed into the network following an interview with a patient.

We generated 50 test cases consisting of ten randomly generated evidence nodes from among the finding nodes defined in the network.[3] For each of the test

---

[3]In addition we conducted tests for different numbers of evidence nodes. Although the performance of our al-

cases, we (1) ran the clustering algorithm on the whole network, (2) ran the relevance-based decomposition algorithm without nuisance node removal, and (3) ran the relevance-based decomposition algorithm with nuisance node removal. In case of the relevance-based decomposition, we selected at each step one target node from among those nodes that had not been updated. We always took the last node on the node list, which was ordered according to the partial order imposed by the graph structure (i.e., parents preceded their children on the list). This procedure gave preference to nodes close to the bottom of the graph. The results of our tests are presented in Figure 6 with the summary data in Table 1. It is apparent that the relevance-based decomposition in combination with the clustering algorithm performed on average over 20 times faster than clustering algorithm applied to the entire network. This difference and the observed variance was small enough to reject possible differences due to chance at $p < 10^{-38}$. Nuisance node removal accounted on the average for over 30% improvement in speed ($p < 10^{-4}$).

|        | Whole netw. | Dec+NuisRem | Dec    |
|--------|-------------|-------------|--------|
| $\mu$  | 464.496     | 32.486      | 24.142 |
| $\sigma$ | 77.720    | 28.728      | 15.927 |
| Min    | 293.190     | 15.320      | 14.250 |
| Median | 471.375     | 20.505      | 17.825 |
| Max    | 632.920     | 181.360     | 98.230 |

Table 1: Summary simulation results for the CPCS network, $n = 50$.

In addition to the CPCS network, we tested the relevance-based decomposition on several other large BBN models. One of these was a randomly generated highly connected network A [Kozlov and Singh, 1996] that we knew was rather difficult to handle for the clustering algorithm. We have not performed tests for fo-cused inference for the A network, as the network was artificial and choosing target nodes randomly would be rather meaningless. Summary results of this test are presented in Table 2. The main reason why standard deviation is larger for the relevance-based decom-gorithm deteriorated as more evidence nodes were added, the algorithm was still faster than belief updating on the entire network even for as many as 40 evidence nodes. We decided to report results for ten evidence nodes, which we believed to be typical for a diagnostic session with CPCS.

|        | Whole netw. | Decomposition |
|--------|-------------|---------------|
| $\mu$  | 213.013     | 20.796        |
| $\sigma$ | 37.074    | 48.430        |
| Min    | 158.750     | 1.283         |
| Median | 203.875     | 7.208         |
| Max    | 305.817     | 331.483       |

Table 2: Summary simulation results for the A network [Kozlov and Singh, 1996], $n = 50$.

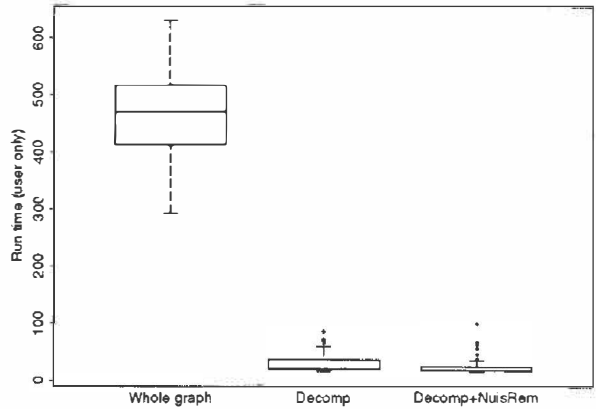

(a)

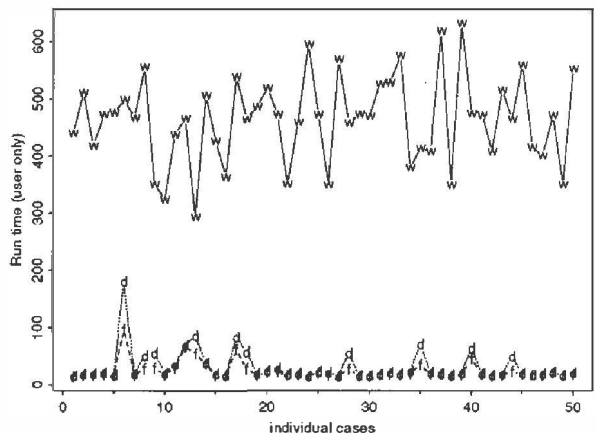

(b)

Figure 6: Comparison of the clustering algorithm applied to the whole network versus the clustering algorithm enhanced with relevance-based decomposition and focused relevance, $n = 50$. Box-plot (a) and time series plot (b) topmost are the times for the whole network, middle for the relevance-based decomposition without nuisance node removal, and bottom, relevance-based decomposition with nuisance node removal.

position algorithm was an outlier of 331.483 seconds. The clustering algorithm took 264.733 seconds for this case. In no other of the 50 cases was the clustering algorithm faster. We also run tests on several networks that we took from a student model of the Andes intelligent tutoring system [Conati et al., 1997] with similar results. Some of the Andes networks were too large to be solved by the clustering algorithm, but were updated successfully by the relevance-based decomposition. Performance differences in case of random tests of tractable Andes networks were minimal and often relevance-based decomposition performed worse than the clustering algorithm applied to the whole network, which confirms that the advantages of relevance-based decomposition are topology-dependent. Focused inference based on relevance reasoning was, on the other hand, consistently orders of magnitude faster than belief updating in the entire network.

One weakness of our experiments that we realized only recently is that we did not have full control over the triangulation algorithm used by the available implementation of the clustering algorithm. We realized that the triangulation algorithm did little in terms of optimizing the size of the junction tree and was sensitive to the initial ordering of the nodes. Relevance algorithm run in the preprocessing phase usually impacted this ordering. Still, we consider it impossible that the observed differences in performance can be attributed to noise in triangulation algorithm — our results are too consistent for this to be a competitive rival hypothesis.

## 6 Discussion

Computational complexity remains a major problem in application of probability theory and decision theory in knowledge-based systems. It is important to develop schemes that will reduce it — even though the worst case will remain NP-hard, many practical cases may become tractable. In this paper, we proposed a computational framework for belief updating in directed probabilistic graphs based on relevance reasoning that aims at reducing the size and connectivity of networks in cases where the inference is focused on a subset of the network's nodes. We introduced relevance-based decomposition, a scheme for computing the marginal distributions of target variables by decomposing the set of target variables into subsets, determining which of the model variables are relevant to those subsets given the new evidence, and performing the computation in the so-identified sub-networks.

As relevance-based decomposition can, for most reasonably sparse topologies, identify sub-networks that are significantly smaller than the entire model. Relevance-based decomposition can also be used to make computation tractable in large networks. A somewhat surprising empirical finding is that this procedure often leads to significant performance improvement even in tractable networks, compared to exact inference in the entire network. One explanation of this finding is that relevance-based techniques are often capable of reducing the clique size at a small computational cost. Roughly speaking, the clustering algorithm constructs a junction tree, whose nodes denote partially overlapping clusters of variables in the original network. Each cluster, or *clique*, encodes the marginal distribution over the set $val(\mathcal{X})$ of the nodes $\mathcal{X}$ in the cluster. The complexity of inference in the junction tree is determined roughly by the size of the largest clique. Reducing the size of the junction tree and breaking large cliques can reduce the complexity of reasoning drastically. Another reason for the observed speedup is that smaller networks are more compatible with the hardware cache on most computer configurations and lead to faster computation by avoiding cache page thrashing. For every computer system, there exist networks that do not fit in its cache or working memory. Decomposition described in this paper will often alleviate possible performance degradation in such cases.

Clustering algorithms aim at distributing the computational complexity between the process of compiling a graph, in which the process of triangularizing the graph is the most important, and belief updating. This is particularly advantageous when a domain model is static in the sense of not being modified while entering evidence and processing probabilistic queries. Methods, as outlined in this paper, seem to be not very suitable for such situations: the framework for relevance reasoning presented in this paper always starts with the initial network and produces reduced networks that need to be compiled from scratch. The cost for using this scheme and all relevance schemes that work on directed graphs is the cost to recompile relevant sub-networks into clique trees before computation. We have found that the relevance algorithms prove themselves worth the cost by sufficient savings in terms of reduced size and connectivity of the network. This can be further enhanced, as one of the reviewers suggested, by caching results of reasoning in overlapping subgraphs. Compilation of and reasoning with the reduced networks may achieve results faster than reasoning with the original network. Application of the proposed schemes suggests that efforts be directed at developing efficient triangularization algorithms that can approach optimality fast and can be used in real-time. Some hope for such schemes has been given in the recent work of Becker and Geiger [1996].

We believe that the relevance-based preprocessing of networks will play a significant role in improving the tractability of probabilistic inference in practical systems. Their computational complexity is low and they can be used as an enhancement to any algorithm, even one that draws significant advantages from precompilation of networks, such as the clustering algorithm used in all test runs in this paper.

## Acknowledgments

This research was supported by the National Science Foundation under Faculty Early Career Development (CAREER) Program, grant IRI–9624629, and by ARPA's Computer Aided Education and Training Initiative under grant number N66001-95-C-8367. We are grateful to Alex Kozlov for making his implementation of the clustering algorithm and several of his benchmark networks available to us. Malcolm Pradhan and Max Henrion of the Institute for Decision Systems Research shared with us the CPCS network with a kind permission from the developers of the Internist system at the University of Pittsburgh. We are indebted to Jaap Suermondt and anonymous reviewers for insightful suggestions.